\documentclass[10pt,twocolumn,letterpaper]{article}

\usepackage{iccv}
\usepackage{times}
\usepackage{epsfig}
\usepackage{graphicx}
\usepackage{amsmath}
\usepackage{amssymb}
\usepackage{newclude}
\usepackage{graphicx}
\usepackage{amsmath}
\usepackage{amssymb}
\usepackage{booktabs}
\usepackage{multirow}
\usepackage{bigstrut}
\usepackage{tabularx}
\usepackage{makecell}
\usepackage{pifont}
\usepackage{breakurl}
\usepackage{color}
\usepackage{stfloats}
\usepackage{graphicx}
\usepackage[accsupp]{axessibility}  

\usepackage[pagebackref=true,breaklinks=true,letterpaper=true,colorlinks,bookmarks=false]{hyperref}
\iccvfinalcopy 

\def\iccvPaperID{3694} 

\ificcvfinal\fi
\begin{document}

\title{Transparent Shape from a Single View Polarization Image}

\author{Mingqi Shao\quad Chongkun\quad Xia Zhendong Yang\quad Junnan Huang\quad Xueqian Wang\thanks{Corresponding author}\\
Tsinghua Shenzhen International Graduate School\\
{\tt\small {\{smq21\, yangzd21, hjn21}@mails.tsinghua.edu.cn\quad \{xiachongkun, wang.xq\}@sz.tsinghua.edu.cn}
}

\maketitle
\ificcvfinal\thispagestyle{empty}\fi

\begin{abstract}
	This paper presents a learning-based method for transparent surface estimation from a single view polarization image. Existing shape from polarization(SfP) methods have the difficulty in estimating transparent shape since the inherent transmission interference heavily reduces the reliability of physics-based prior. To address this challenge, we propose the concept of physics-based prior confidence, which is inspired by the characteristic that the transmission component in the polarization image has more noise than reflection. The confidence is used to determine the contribution of the interfered physics-based prior. Then, we build a network(TransSfP) with multi-branch architecture to avoid the destruction of relationships between different hierarchical inputs. To train and test our method, we construct a dataset for transparent shape from polarization with paired polarization images and ground-truth normal maps. Extensive experiments and comparisons demonstrate the superior accuracy of our method. Our cdataset and code are publicly available at \href{https://github.com/shaomq2187/TransSfP}{https://github.com/shaomq2187/TransSfP}

\end{abstract}


\section{Introduction}
Surface normals provide detailed 3D information about the surface of objects. However, estimating the high-quality surface normal of transparent objects is still an open challenge. The complex rays interactions in transparent objects lead to the difficulty of standard 3D sensors in acquiring accurate surface information\cite{sajjan2020clear}. Besides, transparent objects lack texture of their own, adopting instead the texture of the background, making the regular image-based methods extremely ill-posed and can not produce satisfied estimation\cite{li2020through}. Compared with regular 3D and RGB sensors, the polarization sensor can acquire information about the surface normal from the reflected light on transparent surface. In particular, the commercial polarization image sensors\cite{ohta2017smart} that appeared in recent years have made it possible to capture polarization images at a single shot, making the acquisition of polarization images as easy as conventional sensors. Hence, in this paper, we focus on estimating the surface normals of transparent objects from a single-view polarization image.
\begin{figure}[t]
\centering
\includegraphics[width=\linewidth]{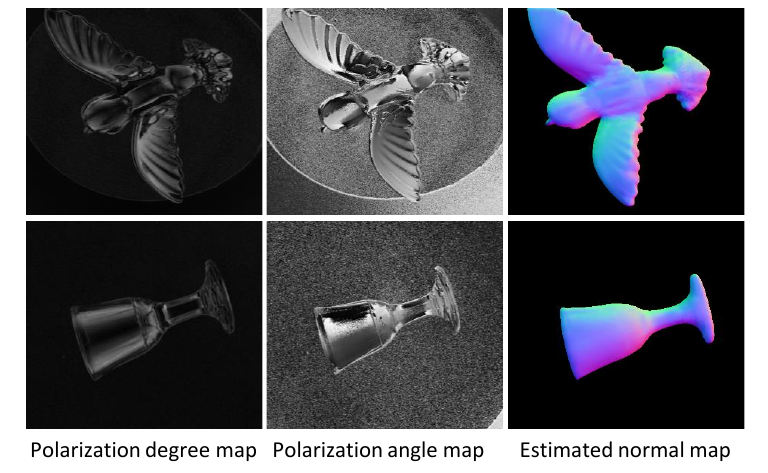}
\caption{\textbf{Examples of our estimation.} Our method is capable of estimating the normal map of various transparent surface from a single view polarization image}
\label{fig:intro}
\end{figure}

The key barrier to using polarization information to estimate the surface normals of transparent objects is the existence of transmission interference. Fresnel specular reflection model(will be described in Sec.\ref{polar_formation}) is commonly used to solve the possible surface normals, i.e. the physics-based prior in SfP, from the observed polarization state. However, due to the high transmittance of transparent objects, the specular reflection energy only accounts for about $4\%$ of the incident light energy\cite{vedel20103d}. The interference transmitted from the background is superimposed on the observed polarization information, leading to the high estimation error of the physics-based SfP methods. 

Aiming at the problem that the reliability of the Fresnel specular reflection model is reduced, we find that the polarization observation of the transmission component has more noise. This characteristic is helpful to distinguish the interfered areas and motivates us to propose the concept of physics-based prior confidence. The confidence indicates the reliability of the physics-based prior and hence serves as the weight of physics-based prior and polarization angle loss. The polarization angle loss is designed to force our network to learn the prior knowledge defined by Fresnel specular reflection model. In addition, we notice that previous learning-based methods\cite{ba2020deep,deschaintre2021deep} directly concatenating physics-based prior and raw polarization images will destroy the relationship between different hierarchies and lead to a decline in the performance. Therefore, we build a network called \textit{TransSfP} with multi-branch architecture to handle different hierarchical inputs. To demonstrate the performance of our method, we construct a dataset for transparent shape polarization, which consists of both real-world and synthetic collections. Benefiting from the above designs, our method can estimate normal maps of various transparent surfacess with lower error as shown in {Fig.\ref{fig:intro}}. We summarize our contributions as: 
\begin{enumerate}
	\item[$\bullet$] We propose a novel learning-based method for transparent surface normal estimation from a single view polarization image.
	\item[$\bullet$] We contribute the first dataset for transparent shape from polarization, which consists of both real-world and synthetic collections.
    \item[$\bullet$] Technically, we introduce three novel designs for transparent SfP problem: the concept of physics-based prior confidence, a multi-branch architecture, and a self-supervised polarization angle loss.

\end{enumerate}


\section{Related Work}
\subsection{Shape from Polarization}
When a beam of unpolarized incident light is reflected from the surface of an object, the polarization state of the light will change according to the incident angle, the relative refractive index, and the surface normal vector of the surface. Early works relied on the physical model of Fresnel equations to recover surface normals. Due to the inherent ambiguity of SfP, other constraints were required to determine a unique surface normal, including rotation measurements\cite{miyazaki2002rotate}, active lighting\cite{activeLighting3}, boundary prior\cite{miyazaki2003polarization} and shading cues\cite{mahmoud2012direct}, but these methods are limited due to strict assumptions and calibrations. With the vigorous development of deep learning in many areas of computer vision, the combination of SfP and deep learning has attracted the attention of researchers. Ba et al.\cite{ba2020deep} first proposed using deep learning to solve the SfP problem and established a real-world object-level dataset. Some works extended `Sfp + deep learning' to other tasks, such as scene-level normal estimation\cite{lei2022shape} and human body shape estimation\cite{zou2022human}. However, these approaches and datasets all focus on opaque objects. As far as we know, there are no publicly available polarization datasets and deep learning models for transparent objects. To address this issue, we propose the first dataset and deep learning model for transparent shape from polarization.
\subsection{Shape Estimation for Transparent Objects}
Shape estimation for transparent objects is a challenging problem\cite{wu2018full}. Active lighting and capture devices are used in many methods to estimate the shape of transparent objects, such as shape from the distortion of the calibration pattern by transparent objects\cite{tanaka2016recovering}, shape from the reflection of transparent objects to known ambient light\cite{yeung2011adequate}, shape from the corresponding relationship between the incident and reflected light\cite{wu2018full}. These methods all require expensive experimental equipment and tedious reconstruction steps. Some researchers tried to use cheaper devices to estimate the surface shape of transparent objects. Li et al.\cite{li2020through} proposed a physics-based neural network, which can reconstruct 3D models of transparent objects from multi-view RGB images and achieve state-of-the-art results. These methods exploited multi-view images to achieve charming results, while our method aims to recover transparent shape from a single view image. There are some works focusing on RGB-D sensor depth completion\cite{sajjan2020clear,zhu2021rgb}, which can complete the depth map of scenes containing transparent objects from a single view. But these works usually focused on transparent object grasping, their estimation accuracy is limited.

Shape from polarization has long been used to estimate the surface of transparent objects with the advantages of single-view,  weak assumption of lighting, and passive imaging. Saito et al.\cite{saito1999measurement} applied Fresnels's law to solve surface orientation by measuring the polarization information of the transparent object in an optical diffuser. To solve the ambiguity in SfP, Miyazaki et al.\cite{miyazaki2004transparent} proposed a method of rotating the transparent object with a tiny angle and matching the feature points of the two polarization states. Since the polarization state of transparent objects is seriously affected by internal reflection, refraction, and transmission, Miyazaki et al.\cite{miyazaki2005inverse} employed inverse raytracing to optimize transparent shapes. However, prior works only focused on simple transparent objects such as spheres and plates. We hope to estimate surface shape of more complex transparent objects by combining SfP with deep learning.


\section{Method}
\subsection{Overview}
\begin{figure*}[tbp]
\centering
\includegraphics[width=0.85\linewidth]{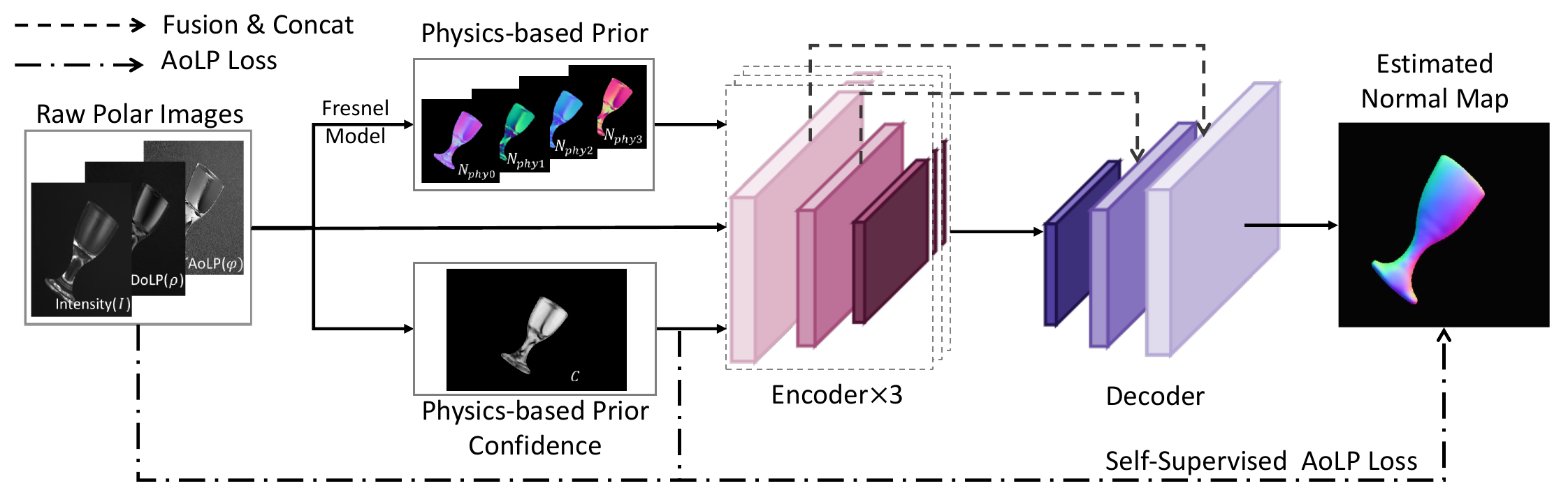}
\caption{\textbf{Overview of our proposed method.} We use a multi-branch architecture to handle different hierarchical inputs rather than directly concatenating them. The physics-based prior confidence map is defined from raw polarization images and used for the weighted fusion of different inputs. In addition, we propose the self-supervised AoLP loss by exploiting the confidence map to force the network to learn the prior knowledge between the normal and AoLP}
\label{fig:overview}
\end{figure*}

Our goal is to estimate transparent surface normal from a single view polarization image. To this end, we present a data-driven approach as shown in Fig.\ref{fig:overview}. We input the raw polarization images to our network, which has three components: intensity map $I$, degree of linear polarization(DoLP, $\rho$) map, and angle of linear polarization(AoLP, $\varphi$) map. Then, we compute the physics-based prior, the four normal maps $N_{phy0}, N_{phy1}, N_{phy2}, N_{phy3}$, by exploiting Fresnel's specular reflection model(Sec.\ref{polar_formation})) and they are input into our network. To handle the transmission interference, we analyze the polarization observation characteristic on transparent surface and define the concept of the physics-based prior confidence $C$ as an additional input(Sec.\ref{confidence}). In addition, we propose a self-supervised AoLP loss based on the confidence to force the network to learn the prior knowledge between AoLP and surface normal(Sec.\ref{network_and_optim}). Finally, we propose separately feeding the different three inputs into the network and the estimated surface normal $\mathbf{\hat{n}}$ can be expressed as:
 \begin{align}
 \mathbf{\hat{n}} = f({\rho,\varphi};{N_{phy0},N_{phy1},N_{phy2},N_{phy_3}};{C})
 \end{align}
 where $f(\cdot)$ represents the prediction model(Sec.\ref{network_and_optim}). In our method, we used the following assumptions: (a) Transparent surface is
smooth and its refractive index is known. (b) The noise in
AoLP map stems from the background, and other sources
are ignored. (c) The reflection component is dominant on
most areas of the transparent surface.

\subsection{Polarization Image Formation of Transparent Shapes}\label{polar_formation}
The ray coming from the transparent surface is the superposition of the reflection component($I_r, \rho_r, \varphi_r$) and the transmission component($I_t, \rho_t, \varphi_t$). Hence, the intensity of a single pixel under a polarizer with an angle of $\alpha_{pol}$ can be expressed as follows:
\begin{align}
I(\alpha_{pol})=&I_{r}[1+\rho_r\cos(2\alpha_{pol}-2\varphi_r)]+ \label{eq:I}\\
&I_{t}[1+\rho_t\cos(2\alpha_{pol}-2\varphi_t)] \nonumber \\ 
=&I[1+\rho\cos(2\alpha_{pol}-2\varphi)]\label{eq:I2}
\end{align}
According to Fresnel's equations, the DoLP and AoLP in the specular reflection component have the following relationship with the zenith $\theta$ and azimuth angle $\phi$ of surface normal $\mathbf{n}=(\sin\theta\cos\phi,\sin\theta\sin\phi,\cos\theta)$:
\begin{align}
\rho_r = \frac{2\sin^2\theta \cos\theta \sqrt{\eta^2-\sin^2\theta}}{\eta^2-\sin^2\theta -\eta^2\sin^2\theta + 2\sin^4\theta}
\label{eq:dolp}
\end{align}
\begin{align}
\varphi_r = \phi \pm \frac{\pi}{2}
\label{eq:aolp}
\end{align}
where $\eta$ is the refraction index and is set to $1.52$ in this paper. Eq.\ref{eq:dolp} and Eq.\ref{eq:aolp} show that two zenith and two azimuth angles can be determined for a given DoLP and AoLP. Four normal maps, i.e. the physics-based prior, can be calculated from the raw polarization images. For detailed calculations of physics-based prior please refer to our supplementary material.

Since the polarization state of the transmission component is unknown, the acquisition of the physics-based prior implies the following approximations: $\rho\approx\rho_r$ and $\varphi \approx \varphi_r$, where $\rho$ and $\varphi$ are the captured DoLP and AoLP. The error of physics-based prior depends on the rationality of the approximations. To understand the impact of the transmission component, we solve $\rho$ and $\varphi$ and introduce the transmission coefficient $T$ to strip the variables in $I_r$ and $I_t$:

\begin{equation}
\resizebox{\linewidth}{!}{$\rho =\frac{\sqrt{I_{r0}^2(1-T)^2\rho_r^2 + I_{t0}^2T^2\rho_t^2+2I_{r0}I_{t0}\rho_r\rho_tT(1-T)\cos(2\varphi_r-2\varphi_t)}}{I_{r0}(1-T)+I_{t0}T}
\label{eq:rho}$}
\end{equation}

\begin{align}
\varphi&=\frac{1}{2}\arctan\frac{I_{r0}\rho_r(1-T)\sin2\varphi_r+I_{t0}\rho_tT\sin2\varphi_t}{I_{r0}\rho_r(1-T)\cos2\varphi_r+I_{t0}\rho_tT\cos2\varphi_t}
\label{eq:variphi}
\end{align}
where $I_{r0}$ and $I_{t0}$ can be regarded as constant values and satisfy the following relationship with $I_r$ and $I_t$: $I_r = (1-T)I_{r0}$, $I_t = TI_{t0}$. Eq.\ref{eq:rho} and Eq.\ref{eq:variphi} illustrate the formation of the polarization state on the transparent surface. The $\rho_t$ is usually small if the background is diffuse and the $\rho_r \gg \rho_t$ since the specular reflection is highly polarized\cite{kalra2020deep}. Hence, the interference of transmission on the DoLP $\rho$ is weak.

Our experimental setting can ensure $\frac{I_{r0}}{I_{t0}}\approx10$. Therefore, according to Eq.\ref{eq:variphi}, the transmission term interference is significant only when $T$ is very close to $1$. In other words, $\varphi$ will be disturbed by $\varphi_t$ heavily when $T\rightarrow1$; otherwise, $\varphi \approx \varphi_r$ is satisfied. This characteristic inspires us to propose the physics-based prior confidence and the AoLP loss.

\subsection{Physics-based Prior Confidence}\label{confidence}

 Using the physics-based prior in the areas with high transmittance will lead to a significant angular error. Although we can not directly obtain the transmittance of each point, fortunately, the characteristic of the transmission component's AoLP $\varphi_t$  can help us to distinguish the areas with strong transmission interference.
\begin{figure}[htbp]
\centering
\includegraphics[width=0.95\linewidth]{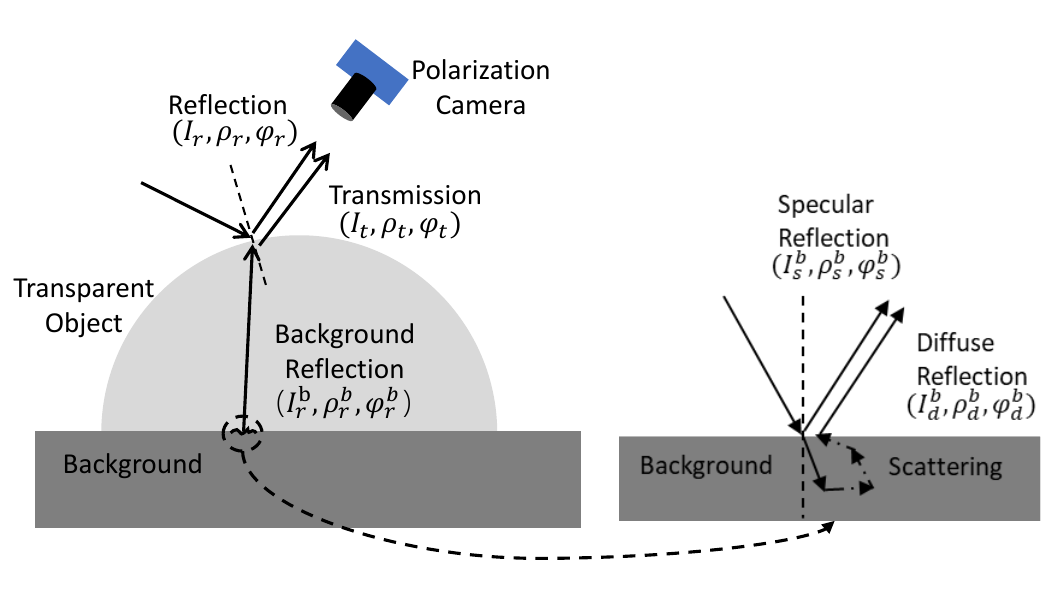}
\caption{\textbf{Different components captured by polarization camera.} The captured ray consists of reflection and transmission components, where the transmission component is composed of two kinds of reflections from the rough background}
\label{fig:reflection}
\end{figure}

\begin{figure}[h]
\centering
\includegraphics[width=1.\linewidth]{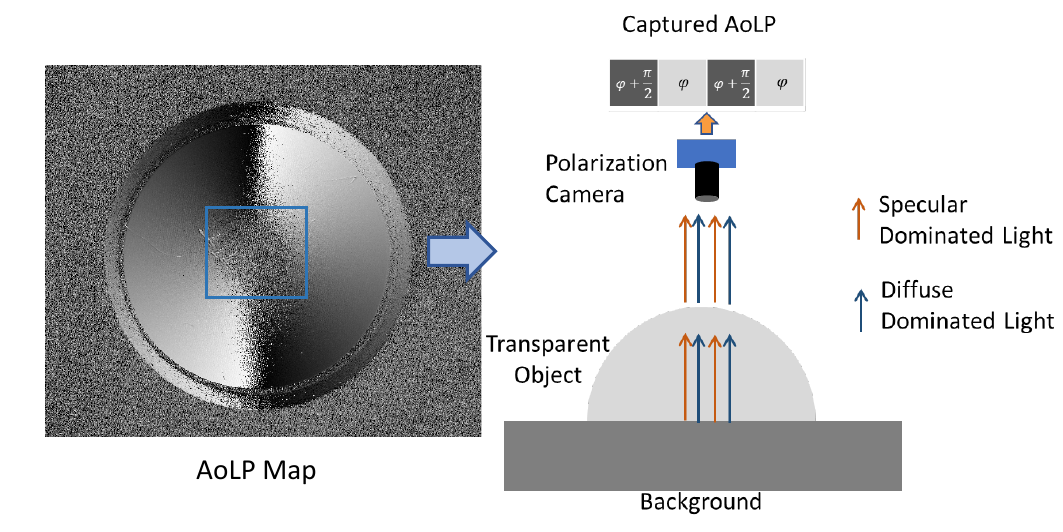}
\caption{\textbf{The characteristic in the AoLP map of the transparent surface.} The noise in the AoLP map is derived from the transition between specular dominance and diffuse dominance on the background surface. More noise indicates higher transmittance and also lower reliability of physics-based prior} 
\label{fig:aolp_fault}
\end{figure}

Compared with the smooth transparent surface, the rays reflected from the rough background $I_r^b$ contain specular $I_s^b$ and diffuse reflection component $I_d^b$ as shown in Fig.\ref{fig:reflection}, which can be expressed as follows:
\begin{align}
I_r^b(\alpha_{pol})=&I_{s}^b[1+\rho_s^b\cos(2\alpha_{pol}-2\varphi_s^b)]+ \\
&I_{d}^b[1+\rho_d^b\cos(2\alpha_{pol}-2\varphi_d^b)] \nonumber \\ 
=&I_r^b[1+\rho_r^b\cos(2\alpha_{pol}-2\varphi_r^b)]
\end{align}
The shape from polarization theory gives us that the AoLPs of specular reflection and diffuse reflection are orthogonal\cite{atkinson2006recovery}: 
\begin{align}
\varphi_s^b = \varphi_d^b \pm \frac{\pi}{2}
\end{align}
Then, we can solve for $\varphi_t$ in terms of $I_s^b, \rho_s^b, \varphi_s^b, I_d^b,\rho_d^b$:
 \begin{align}
 \varphi_r^b=\varphi_s^b-\frac{1-{\mathrm{sign}}(
 I_s^b\rho_s^b-I_d^b\rho_d^b)}{2}\pi
 \label{eq:phi_t}
 \end{align}
It can be seen that there is a sign function in the $\varphi_r^b$ expression, which will lead to the discontinuity in the AoLP map. For the rough background, both specular and diffuse reflection exists and their dominance can not be determined, resulting in the noise in AoLP map. In fact, the polarization state of $I_r^b$ will change according to the incidence angle $\theta_i$ of the interaction point when the ray passes through the transparent object. We denote this process as $\mathrm{M_t(\cdot)}$ and the following expression can be obtained:
 \begin{align}
 I_t(\alpha_{pol})=\mathrm{M_t}(I_r^b(\alpha_{pol});\theta_i)
 \end{align}
 The incident angles of rays from a local area are approximately equal, which means that they will gain the same phase shift from $\mathrm{M_t(\cdot)}$. Therefore, the noise in $\varphi_r^b$ also exists in $\varphi_t$. According to the above analysis, the AoLP observation of transparent surface has the characteristic that the area with higher transmittance contains more noise as shown in Fig.\ref{fig:aolp_fault}. 

Above analysis illustrates that the noise in the AoLP map is related to the transmittance and physics-based prior reliability. Therefore, we define the concept \textbf{Physics-based Prior Confidence} by quantifying the noise in the AoLP map. The noise for a given pixel can be quantified by the distance between the pixel and its neighborhood pixels. We first define the distance $d_{K,m}(i,j)$ between the pixels in the $K\times K$ neighborhood of point $(i,j)$:
 \begin{align}
 d_{K,m}(i,j) = \sum_{p\in P_{i,j}} |p-\bar{p}_{i,j}|^m 
 \label{eq:noise_density}
 \end{align}
where $P_{i,j}$ represents the set of pixel values in the $K\times K$ neighborhood of point $(i,j)$, $\bar{p}_{i,j}$ is the mean of pixel values belonging to the set $P_{i,j}$ and $m$ is the smoothing exponential term. $K$ determines the consistency between the AoLP map and confidence map and its value in this paper is set to $9$. The parameter $m$ controls the mapping relationship between noise density and the value of confidence and its default value is $0.5$. By normalizing the distance map, the physics-based prior confidence can be obtained:
 \begin{align}
 C_{K,m}(i,j) = 1-\frac{d_{K,m}(i,j)}{\max \limits_{\substack{0\leq x<W, 0\leq y<H}} d_{K,m}(x,y)}
 \label{eq:confidence}
 \end{align}
where $H, W$ denote the height and width of the AoLP map respectively. The physics-based prior confidence defined by the above equation is also low at the junction of $0$ and $\pi$ in the AoLP map (according to Eq.\ref{eq:I2}, the value range of AoLP is $[0,\pi]$). This situation is allowed because the lower confidence of the physics-based prior in such area is helpful to avoid the network being misguided by the jumped physics-based prior(see additional results which are provided in our supplementary material).


We first input the physics-based prior confidence map as an addition prior into the network and guide the fusion of raw polar images and physics-based prior to minimize prediction error. Then it is also used as the weight of the self-supervised AoLP loss.

\subsection{Network and Optimization}\label{network_and_optim}


\begin{figure*}[t]
\centering
\includegraphics[width=\linewidth]{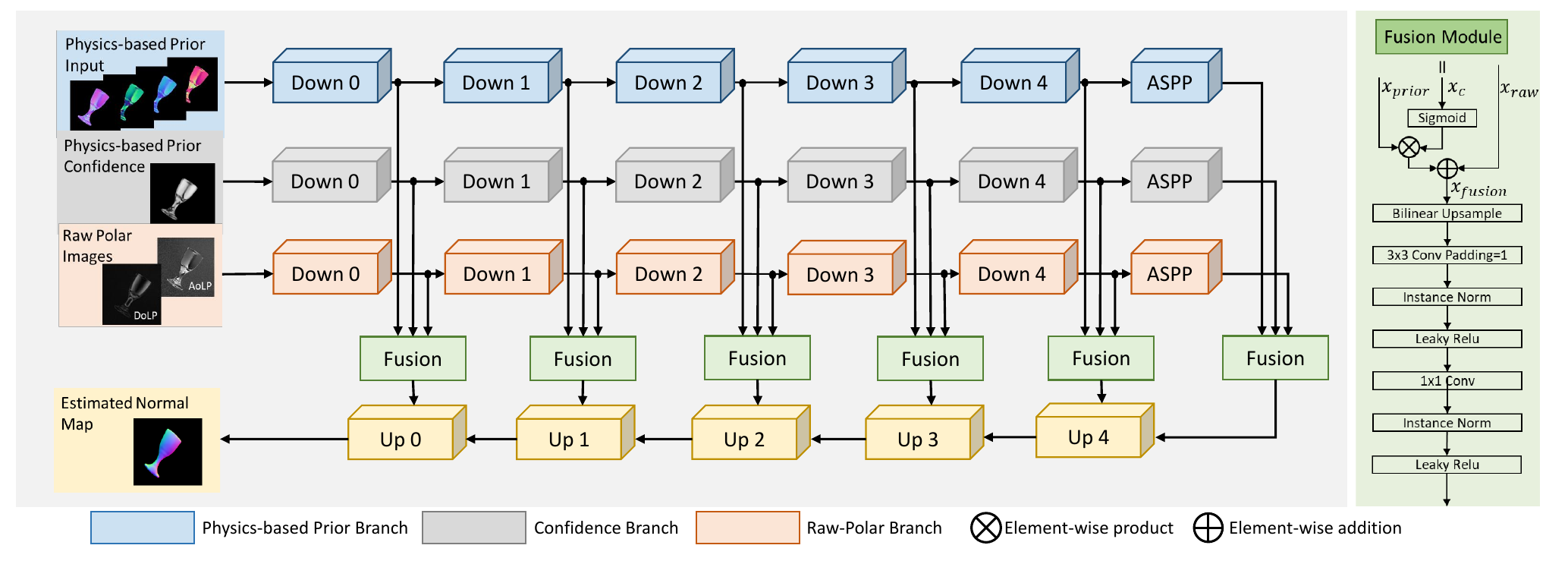}
\caption{\textbf{Overview of TransSfP.} We regard raw polar images, physics-based prior, and confidence map as different hierarchical information and input them into three independent encoders. The confidence branch is used for weighting the features from physics-based prior branch, which is completed in our proposed fusion module}
\label{fig:architecture}
\end{figure*}

A straightforward way to use deep learning to solve the SfP problem is to concatenate the raw polarization images and physics-based prior into the network for normal estimation\cite{ba2020deep,lei2022shape}. Different with previous methods, we regard transparent SfP as a multi-modal fusion problem, using separate encoders to extract features of the different inputs. Features from the raw polar branch and physics-based prior branch are fused at multiple levels according to the output of the confidence branch, and finally, a decoder is employed to complete the normal estimation.

 As shown in {Fig.\ref{fig:architecture}}, our network \textit{TransSfP} consists of three encoders and one decoder. The encoders are the physical prior branch, the physics-based prior confidence branch, and the raw polarization branch respectively. The structures of the encoders are completely the same, but the parameters are independent of each other. EPSANet50(Small) is adopted as the backbone of the Encoder, which is an improved backbone that replaces the Bottleneck in ResNet50 by ESPANet Block which exploits channel attention mechanism\cite{zhang2021epsanet}. At the beginning of the Encoder, the module named `Down 0' consisting of a $1\times1$ convolutional layer and a BatchNorm layer is used to force the network to extract the per-pixel features. At the end of the Encoder, we add an ASPP(Atrous Spatial Pyramid Pooling) module to enhance the network's ability to capture multi-scale information\cite{chen2018encoder}. The detailed structure of each module is provided in the supplementary material.
 
 To mitigate the adverse contribution of interfered physics-based prior, we propose a fusion module as shown in the right of  Fig.\ref{fig:architecture}. Features from the confidence branch $x_c$ are first converted to weights in the range $(0,1)$ by passing through a Sigmoid function and then multiply with features from the physics-based prior branch $x_{prior}$ to control the contribution of physics-based prior. Then, the U-Net style skip connection\cite{ronneberger2015u} is employed to introduce the fused features of different levels into the upsampling modules. 
  


Lastly, we adopt the following loss function to optimize our network:
 \begin{align}
 \mathcal{L}_{net} = \mathcal{L}_{sim} + \lambda\mathcal{L}_{aolp}
 \label{eq:L_net}
 \end{align}
where $\mathcal{L}_{sim}$ is the cosine similarity loss and $\mathcal{L}_{aolp}$ is the AoLP loss. The default value of $\lambda$ is set to $0.05$. 

The cosine similarity loss is commonly used in normal estimation, and its expression is as follows:
 \begin{align}
 \mathcal{L}_{sim} = 	\sum_{i=0}^{W}\sum_{j=0}^{H}(1-\frac{\mathbf{n}_{i,j}\cdot\mathbf{\hat{n}}_{i,j}}{\Vert \mathbf{n}_{i,j}\Vert_2 \ \Vert \mathbf{\hat{n}}_{i,j}\Vert_2 })
 \label{eq:similarity}
 \end{align}
  where $\mathbf{n}_{i,j}$ represents the ground-truth normal vector at the point $(i,j)$, and $\mathbf{\hat{n}}_{i,j}$ is the estimated normal vector.
  
  We propose the AoLP loss to force the network to learn the physical knowledge represented by Eq.\ref{eq:aolp}. Due to the existence of transmission interference, Eq.\ref{eq:aolp} can not be accurately satisfied. Hence, we take the physics-based prior confidence as the weight of the AoLP error to mitigate this interference:
  \begin{align}
  \mathcal{L}_{aolp} = 	\sum_{i=0}^{W}\sum_{j=0}^{H}c_{i,j}\min(\vert\varphi_{i,j}+\frac{\pi}{2}-\hat{\phi}_{i,j} \vert,\vert\varphi_{i,j}-\frac{\pi}{2}-\hat{\phi}_{i,j} \vert)
  \label{eq:aolp_loss}
  \end{align}
  where $c_{i,j}$ is the value of point $(i,j)$ in the confidence map and $\hat{\phi}_{i,j}$ represents the azimuth of $\mathbf{\hat{n}}_{i,j}$. $\mathcal{L}_{aolp}$ is a self-supervised loss term since only AoLP and confidence map are used in the computation.
	
\subsection{Dataset}

We establish a dataset for transparent shape from polarization, which contains raw polarization images, physics-based prior(four normal maps), ground-truth masks and normal maps. Compared with prior polarization datasets\cite{ba2020deep,lei2022shape,kondo2020accurate}, our dataset is the first dataset for transparent shape from polarization.

\begin{figure}[htbp]
\centering
\includegraphics[width=\linewidth]{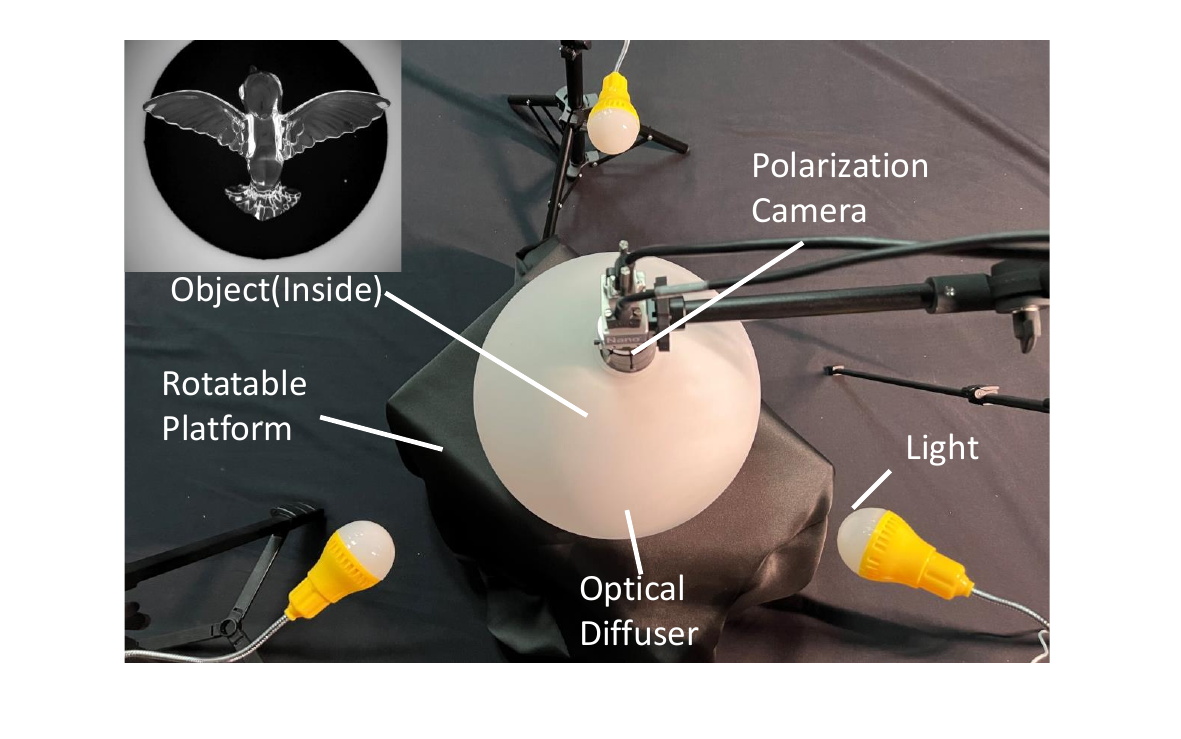}
\caption{\textbf{Setup for acquiring real-world dataset}}
\label{fig:setup}
\end{figure}
 
Our dataset consists of two parts: the real-world dataset and the synthetic dataset. {Fig.\ref{fig:setup}} is our real-world setup for acquiring our dataset. As discussed before, the polarization information of the transmission component will bring adverse effect to normal estimation, hence, our dataset acquisition setup should increase the reflection and reduce the transmission as much as possible. An optical diffuser made of frosted glass with diameter of $30 cm$ is used to simulate global illumination so that the transparent surface has strong and uniform reflections in all directions. To reduce the transmission from the background, the transparent object is placed on rough background. A DLASA G3-GM14-M2450 polarization camera is employed as our capture device, which can capture four polarization images(at $\theta_{pol}$ of $0^\circ, 45^\circ, 90^\circ, 135^\circ$) at a single shot. We coate transparent objects with powder and reconstructed their 3D models by using a 3D scanner with an accuracy of $0.1 mm$, and then manually align them with the captured polarization images in the unity game engine to obtain the ground-truth mask and normal maps. Using the above steps, we build the real-world dataset of $10$ different objects, each of which is rotated between $0^\circ$ and $360^\circ$ at $5^\circ$-$10^\circ$ intervals, resulting in a total of $486$ samples. 

The synthetic dataset is created since the real-world data acquisition requires manual alignment, resulting in the difficulty of acquiring enough real-world data for training. We establish a scene similar to the setup in {Fig.\ref{fig:setup}} and implement an integrator capable of outputting a camera-space normal map in the mitsuba2\cite{nimier2019mitsuba}, a physically-based render that can track the full polarization state of light during a simulation. To simulate the diffuser's uniform lighting, we attach the diffuser's surface with an area light source. The background surface uses the diffuse BSDF with the reflectance of $0.1$, and the transparent object adopts the dielectric BSDF with the refractive index of $1.52$. The 3D models of $13$ objects, collected from DiLiGenT-MV dataset\cite{li2020multi} and Stanford 3D Scanning Repository\cite{levoy2005stanford}, are placed in the scene we build to run polarization rendering. Each object is rotated $72$ times from $0^\circ$ to $360^\circ$, resulting in a total of $936$ samples in the synthetic dataset.

\subsection{Implementation Details}
We implement our model on PyTorch and the model is trained on an NVIDIA GeForce RTX 3090 GPU(24GB) with a batch size of 5. The Adam optimizer\cite{kingma2014adam} with an initial learning rate of 1e-6 and a weight decay coefficient of 5e-4 is used for optimizing the network. We adopt the Step learning rate decreasing strategy, the learning rate times $0.1$ every 9 epochs. To reduce memory usage, we crop the images to $512\times 512$ patches in the data augmentation stage.

%



\section{Experiments}
\subsection{Experimental Setup}
We use the evaluation metrics commonly used in normal estimation, including mean angular error(\textit{mean, MAE}), median angular error(\textit{median}), and \textit{Accuracy} ${\it 11.25^\circ}$, $\it 22.5^\circ$, $\it 30^\circ$, which represent the ratio of the number of pixels with an error lower than this value to the total number of valid pixels. All samples in the synthetic dataset are used for training, and some objects in the real-world dataset are also employed for training to make up for the difference between the synthetic and the real-world collections. 
\begin{figure*}[htbp]
\centering
\includegraphics[width=0.95\linewidth]{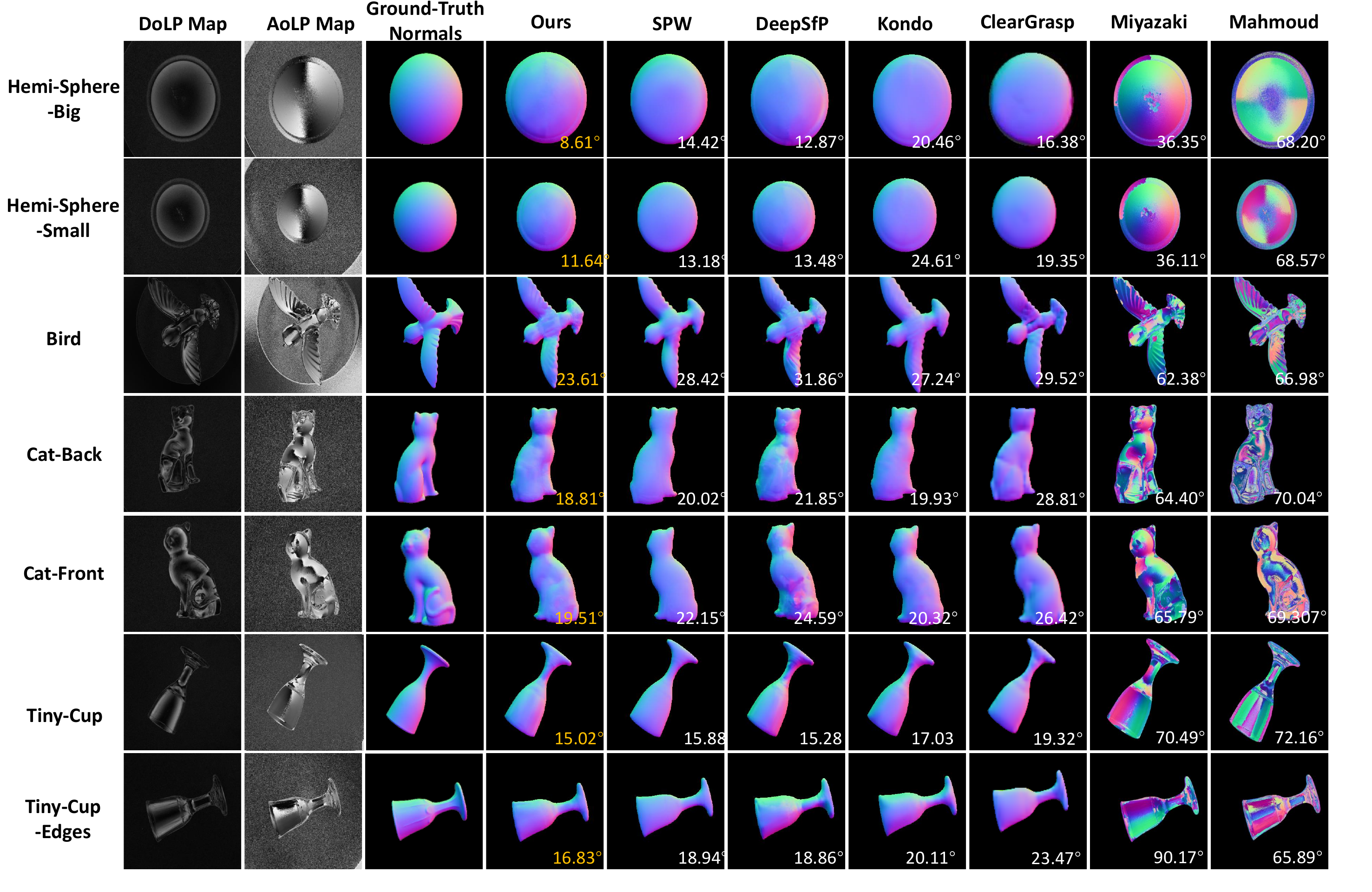}
\caption{\textbf{Visual results of the comparisons to baselines.} Our method outperforms all baselines on all test objects}
\label{fig:baselines}
\end{figure*}
\subsection{Comparisons to Baselines}\label{sec:baseline}
\begin{table*}[htbp]
  \centering
  \caption{\textbf{Quantitaive results of the comparisons to baselines.} The best results in the table are marked in \textbf{Bold}}
    \begin{tabular}{ccccccccc}
    \toprule
    \multirow{2}[4]{*}{Method} & \multicolumn{8}{c}{Mean Angular Error $\downarrow$} \\
\cmidrule{2-9}          & \makecell[c]{Hemi-Sph\\-Big} & \makecell[c]{Hemi-Sph\\-Small} & Bird  & Cat-Back & Cat-Front & Tiny-Cup & \makecell[c]{Tiny-Cup\\-Edges} & All \\
    \midrule
    Miyazaki\cite{miyazaki2003polarization} & 36.11° & 36.35° & 62.38° & 64.40° & 65.79° & 70.49° & 90.17° & 55.62° \\
    Mahmoud\cite{mahmoud2012direct} & 68.57° & 68.20° & 66.98° & 70.04° & 69.30° & 72.16° & 65.89° & 68.72° \\
    ClearGrasp\cite{sajjan2020clear} & 19.35° & 16.38° & 29.52° & 28.81° & 26.42° & 19.32° & 23.47° & 23.32°\\    
    DeepSfP\cite{ba2020deep} & 13.48° & 12.87° & 31.86° & 21.85° & 24.59° & 15.28° & 18.86° & 19.83° \\
    SPW\cite{lei2022shape}   & 13.18° & 14.42° & 28.42° & 20.02° & 22.15° & 15.88° & 18.94° & 19.00° \\
    Kondo\cite{kondo2020accurate} & 24.61° & 20.46° & 27.24° & 19.93° & 20.32° & 17.03° & 20.11° & 21.39° \\
    \textbf{Ours}  & \textbf{8.61°} & \textbf{11.64°} & \textbf{23.61°} & \textbf{18.81°} & \textbf{19.51°} & \textbf{15.02°} & \textbf{16.83°} & \textbf{16.29°} \\
    \bottomrule
    \end{tabular}%
  \label{tab:baselines}%
\end{table*}%
In this paper, we aim to recover the transparent surface normals from polarization images in a single view. Hence, we employ two physics-based SfP methods, Miyazaki et al.\cite{miyazaki2003polarization}, Mahmoud et al.\cite{mahmoud2012direct}, and three learning-based SfP methods, DeepSfP\cite{ba2020deep},  Kondo et al.\cite{kondo2020accurate}, SfP in the Wild(SPW)\cite{lei2022shape} as the SfP baselines of our method. In addition, ClearGrasp\cite{sajjan2020clear}, whose normal prediction module can estimate the normals of transparent shapes using a single RGB image, is also employed as one of the baselines. We retained the learning based-SfP methods on our dataset. For the ClearGrasp, we performed fine-tuning on our dataset for the model pre-trained on the ClearGrasp dataset. Mahmoud's method needs to provide albedo and light source direction, we assume the albedo is uniform of 1, and the light source direction is estimated using the method proposed by Smith et al.\cite{smith2018height}. 


{Fig.\ref{fig:baselines}} lists the visual results of the comparison with baselines, {Table \ref{tab:baselines}} lists the quantitative results of the comparison. Our method achieves the best performance on all objects in the test set. A notable result is the comparison between ClearGrasp and other learning-based SfP methods. Though ClearGrasp uses large-scale synthesized RGB images for pretraining, its performance is still worse than any other learning method using polarization information, which illustrates the advantages of polarization sensor over regular RGB sensor.


\subsection{Comparisons of Network Architectures}\label{sec:network}

\textit{TransSfP} uses the multi-branch architecture as shown in {Fig.\ref{fig:architecture}}. To illustrate the effectiveness of our architecture, we conduct the comparison with several commonly used semantic segmentations networks such as U-Net\cite{ronneberger2015u}, DeepLabV3+\cite{chen2018encoder} and models for normal estimation: PS-FCN\cite{chen2018ps}, DeepSfP\cite{ba2020deep}, SPW\cite{lei2022shape}, Kondo\cite{kondo2020accurate} and single branch version of \textit{TransSfP}. Our multi-branch architecture inputs the raw-polar input, physics-based prior input, and transmission confidence map separately, while other architectures concatenate them directly.
\begin{table}[htbp]
  \centering
  \caption{\textbf{Comparisons of network architectures}}
  \resizebox{\linewidth}{!}
  {
    \begin{tabular}{cccccc}
    \toprule
    \multirow{2}[2]{*}{Network} & \multicolumn{2}{c}{Angular Error $\downarrow$} & \multicolumn{3}{c}{Accuracy $\uparrow$} \\
          & Mean  & Median & 11.25° & 22.5°  & 30° \\
    \midrule
    U-Net\cite{ronneberger2015u} & 25.20° & 22.90° & 16.06\% & 49.42\% & 69.07\% \\
    PS-FCN\cite{chen2018ps} & 25.93° & 22.42° & 22.06\% & 52.79\% & 71.12\% \\
    DeepLabV3+\cite{chen2018encoder} & 21.90° & 18.93° & 21.04\% & 63.78\% & 81.84\% \\
    DeepSfP\cite{ba2020deep} & 19.00° & 14.91° & 38.36\% & 77.36\% & 87.48\% \\
    SPW\cite{lei2022shape}   & 21.73° & 18.43° & 28.38\% & 61.24\% & 84.00\% \\
    Kondo\cite{kondo2020accurate} & 24.40° & 22.20° & 18.95\% & 54.39\% & 75.09\% \\
    Ours(Single Branch) & 18.73° & 15.28° & 33.72\% & 77.67\% & 88.65\% \\
    \textbf{Ours}  & \textbf{16.29°} & \textbf{12.85°} & \textbf{48.31\%} & \textbf{83.20\%} & \textbf{93.90\%} \\
    \bottomrule
    \end{tabular}%
   }
  \label{tab:architectures}%
\end{table}%

 The raw polarization information and the physics-based prior are converted through Frenel's physical model, hence they can be regarded as different hierarchies of polarization information. Direct concatenation will destroy the relationship between different hierarchies. The quantitative results in Table \ref{tab:architectures} prove that our consideration of treating them as different hierarchical information and then transforming them into a multi-modal fusion problem is effective.


\begin{table}[htbp]
  \centering
  \caption{\textbf{Ablation experiment results.} The module with `*' indicates that the single branch architecture is used}
  \resizebox{\linewidth}{!}
  {
    \begin{tabular}{cccccc}
    \toprule
    \multirow{2}[2]{*}{Studied Module} & \multicolumn{2}{c}{Angular Error $\downarrow$} & \multicolumn{3}{c}{Accuracy $\uparrow$} \\
          & Mean  & Median & 11.25° & 22.5°  & 30° \\
    \midrule
    W/o Polarization* & 24.70° & 22.99° & 18.32\% & 52.84\% & 74.25\% \\
    With Polarization* & 18.73° & 15.29° & 33.72\% & 77.67\% & 91.65\% \\
    W/o Confidence & 17.72° & 14.47° & 40.02\% & 81.08\% & 92.06\% \\
    W/o AoLP Loss & 17.17° & 14.11° & 41.81\% & 82.06\% & 93.14\% \\
    \textbf{Full}   & \textbf{16.29°} & \textbf{12.85°} &\textbf{ 48.31\%} & \textbf{83.20\%} & \textbf{93.90\%} \\

    \bottomrule
    \end{tabular}%
   }
  \label{tab:confidence}%
\end{table}%
 \begin{figure}[htbp]
 \centering
 \includegraphics[width=\linewidth]{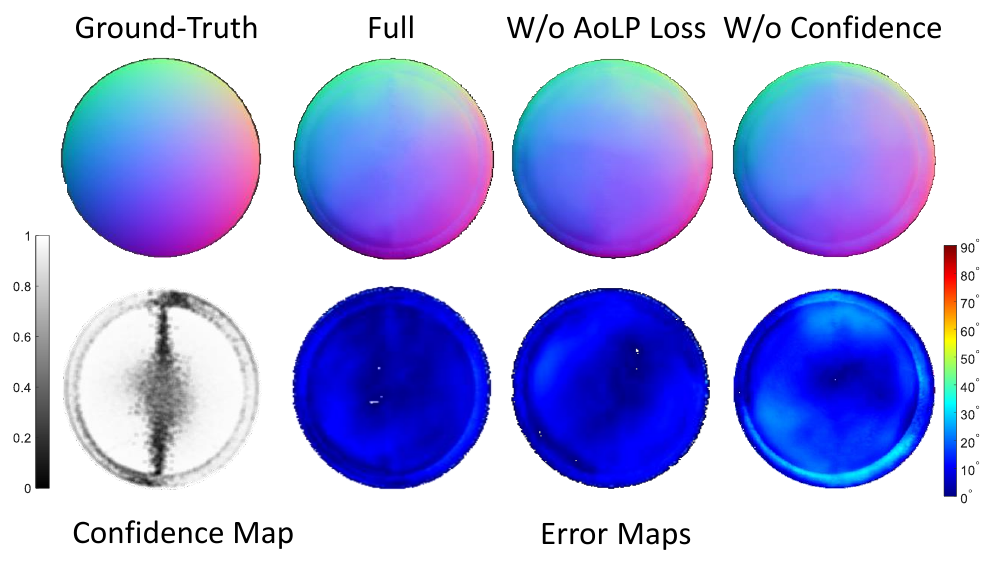}
 \caption{\textbf{Importance of AoLP loss and prior confidence}}
 \label{fig:confidence}
 \end{figure}
\subsection{Ablation Studies}
\textbf{Importance of Polarization.} In Section.\ref{sec:baseline}, the performance of ClearGrasp using RGB images is worse than any learning-based SfP method, which preliminarily illustrates the importance of polarization. To avoid the influence of network architecture, we conducted a further ablation experiment. We adopt the single branch architecture mentioned in Section.\ref{sec:network}, using light intensity and polarization information as input, respectively. The results listed in Table \ref{tab:confidence} show that the performance of using polarization information is much better than using only intensity since polarization cues contain more information about the surface normal. Both the physical insights and the ablation study prove the superiority of the polarization sensor in the surface estimation of transparent objects.

\textbf{Importance of Physics-based Prior Confidence.} We set the confidence to constant $1$ to remove the confidence branch. The quantitative and qualitative results in Fig.\ref{fig:confidence} and Table \ref{tab:confidence} show that our proposed prior confidence is beneficial to transparent shape from polarization. The physics-based prior defined by the Eq.\ref{eq:dolp} and Eq.\ref{eq:aolp} provides rich information for normal estimation while the prior of some areas may have adverse effects due to the transmission interference. The confidence we proposed forces the network more conservative to use the physics-based prior of these areas and hence can reduce the adverse contribution significantly. 


\textbf{Importance of AoLP Loss}. Fig.\ref{fig:confidence} shows that the AoLP loss can reduce the error in areas with high confidence(or weak transmission) because the approximation of $\varphi\approx\phi \pm \frac{\pi}{2}$ is reliable in these areas. The AoLP loss can force the network to learn this prior knowledge and thus improving the prediction accuracy.


\section{Conclusions}
In this paper, we present a novel method for transparent shape from polarization. We demonstrate the performance of our method on our dataset, the first dataset for transparent shape from polarization. By introducing the physics-based prior, multi-branch network architecture, and self-supervised AoLP loss, the negative contribution of the interfered physics-based prior is effectively reduced and our method outperforms previous approaches on all objects in the test set. We hope our dataset and model can contribute to the community of transparent shape estimation.


\noindent\textbf{Limitations} Our method alleviates the transmission interference, however, when the most areas of transparnt surface are dominated by transmission component, our method will degenerate to a rgb-based method. In addition, we notice that the pure polarization information limits the overall accuracy of transparent shape estimation. Combining polarization information with other reconstruction methods is one of our future research directions.
\section*{Acknowledgement}
This work was supported by the National Key R\&D Program of Chin 2022YFB4701400/4701402), National Natural Science Foundation of Chin No. U21 B6002, 62203260, 92248304), Guangdong Basic and Applied Bas esearch Foundation (2023A1515011773), Science, Technology and Innova on Commission of Shenzhen Municipality (WDZC20200820160650001CYJ20200109143003935).

{\small
\bibliographystyle{ieee_fullname}
\bibliography{egbib_iccv}
}

\end{document}